\documentclass[conference]{IEEEtran}
\IEEEoverridecommandlockouts
\usepackage{cite}
\usepackage{amssymb,amsfonts}
\usepackage{algorithm}
\usepackage{algpseudocode}
\usepackage{graphicx}
\usepackage{textcomp}
\usepackage{xcolor}
\usepackage[hidelinks]{hyperref}
\usepackage{amsmath}
\usepackage{mathtools}
\usepackage{subcaption}
\usepackage{amsthm}
\usepackage{bbm}
\usepackage{siunitx}
\usepackage{MnSymbol}
\usepackage{wasysym}
\usepackage{amsfonts}
\usepackage{booktabs}
\usepackage{multirow}

\newtheorem{assumption}{Assumption}[]
\newtheorem{definition}{Definition}[]

\newtheorem{lemma}{Lemma}[]
\newtheorem{thm}{Theorem}[]

\algnewcommand\algorithmicforeach{\textbf{for each}}
\algdef{S}[FOR]{ForEach}[1]{\algorithmicforeach\ #1\ \algorithmicdo}

\newcommand\numberthis{\addtocounter{equation}{1}\tag{\theequation}}

\def\BibTeX{{\rm B\kern-.05em{\sc i\kern-.025em b}\kern-.08em
    T\kern-.1667em\lower.7ex\hbox{E}\kern-.125emX}}
\begin{document}

\title{Fairness-aware Federated Minimax Optimization with Convergence Guarantee
\thanks{This work was supported by a grant from the NSFC/RGC Joint Research Scheme sponsored by the Research Grants Council of the Hong Kong Special Administrative Region, China and National Natural Science Foundation of China (Project No. N\_HKUST656/22).}}

\author{
\IEEEauthorblockN{Gerry Windiarto Mohamad Dunda}
\IEEEauthorblockA{\textit{Electronic and Computer Engineering} \\
\textit{The Hong Kong University of Science and Technology}\\
Clear Water Bay, Hong Kong \\
gwmdunda@connect.ust.hk}
\and
\IEEEauthorblockN{Shenghui Song}
\IEEEauthorblockA{\textit{Electronic and Computer Engineering} \\
\textit{The Hong Kong University of Science and Technology}\\
Clear Water Bay, Hong Kong \\
eeshsong@ust.hk}
}

\maketitle

\begin{abstract}
Federated learning (FL) has garnered considerable attention due to its privacy-preserving feature. Nonetheless, the lack of freedom in managing user data can lead to group fairness issues, where models are biased towards sensitive factors such as race or gender. To tackle this issue, this paper proposes a novel algorithm, fair federated averaging with augmented Lagrangian method (FFALM), designed explicitly to address group fairness issues in FL. Specifically, we impose a fairness constraint on the training objective and solve the minimax reformulation of the constrained optimization problem. Then, we derive the theoretical upper bound for the convergence rate of FFALM. The effectiveness of FFALM in improving fairness is shown empirically on CelebA and UTKFace datasets in the presence of severe statistical heterogeneity.  
\end{abstract}


\section{Introduction}
Federated learning (FL) \cite{FedAvg} is a distributed machine learning approach that enables model training on potentially sensitive data from different entities without the necessity for data sharing. This technique is promising in diverse domains such as computer vision (CV) as it can facilitate training of models on a large-scale, diverse set of data while preserving data privacy. However, a direct implementation of existing federated algorithms may violate group fairness \cite{groupfairness}, which refers to the equitable treatment of different groups in a population. Group fairness is required by law such as in Europe \cite{eulaw}, enforcing that the decision making by predictive models does not exhibit bias towards any particular sensitive group, such as race or gender. For example, an AI model used in a hiring process may have been trained on historical data that reflects biased hiring patterns, leading to discriminatory outcomes for underrepresented groups in the workforce. There are more examples \cite{unfaircvexamples} that further motivate raising awareness in training fair deep learning models.

The sources of group unfairness or bias mainly come from dataset, which may reflect measurement or historical bias from the annotators, and the training algorithm, which may learn unwanted biased features from such dataset. The aforementioned sources also induce statistical heterogeneity. Under severe heterogeneity, the trained model may have poor task performance and biased model. Finding FL algorithms that are fair and robust to statistical heterogeneity or non-identical and independently distributed (non-iid) data is an arduous task, and currently it is an open problem \cite{openproblem}. 

\subsection{Contributions} Since we have little control on the clients data in FL, efforts to mitigate bias from the federated algorithm are important. Therefore, the focus of this paper is to improve group fairness in binary classification tasks, involving binary sensitive attributes, which is common in fairness literature \cite{binaryfair}. In particular, we propose a new federated algorithm that is effective in reducing bias while maintaining similar task performance as existing federated algorithms. The main contributions are summarized below.
\begin{itemize}
    \item We propose a fairness-aware algorithm, fair federated averaging with augmented Lagrangian method (FFALM). Firstly, we formulate a constrained minimization problem on the global loss function satisfying a fairness metric. Inspired by augmented Lagrangian method \cite{numericalopt}, we solve the problem by leveraging the local training as a sub-solver to find the optimal model parameters given dual iterates. Then, the dual iterates are locally updated given the updated model parameter, and they are aggregated using weighted average. 
    \item We propose a theoretical upper bound for the convergence rate of FFALM over a nonconvex-strongly-concave objective function, which is $\mathcal O(\frac 1 {T^{2/3}})$.
    \item We empirically assess the proposed method on publicly accessible CV datasets (CelebA and UTKFace) containing sensitive attributes (gender and skin color respectively) based on two common fairness metrics: demographic parity difference (DPD) and equal opportunity difference (EOD). Experimental results show that FFALM improves DPD by $10 \%$ and EOD by $14 \%$ in the attractiveness prediction task, and DPD by $6 \%$ and EOD by $5 \%$ in the youth prediction task compared to FedAvg under severe data heterogeneity.
\end{itemize}

\section{Related Work}
There have been some engaging results in tackling the fairness issues in deep-learning models. In the following, we categorize some prior related works based on how the training is conducted, either centralized or federated.

\subsection{Ensuring fairness in centralized learning} In centralized learning, it is not uncommon to modify the training framework to achieve a suitable degree of group fairness. The authors of \cite{balancednotenough} decorrelated the input images and the protected group attributes by using adversarial training. Knowledge transfer techniques and multi-classifiers can also be adopted as a debiasing method \cite{inclusivefacenet}. Augmenting each image sample with its perturbed version generated from generative models can potentially reduce biases as well \cite{augmentdebias}. The aforementioned works require additional components to the model, thus increasing the computation cost. This might not be suitable for FL. A possible alternative is to alter the loss function to take into account group fairness. The authors of \cite{FairALM} introduced a loss function obtained from the upper bound of the Lagrangian based on a constrained optimization formulation from a finite hypothesis space perspective.

\subsection{Ensuring fairness in FL} Some prior works considered group fairness in FL. Due to system constraints, most innovations came from the objective formulation to include fairness or more information exchange between the server and the clients. The example for the latter is FairFed \cite{fairfed}, where the client coefficients are adaptively adjusted during the aggregation phase based on the deviation of each client's fairness metric from the global average. Along the line of the objective formulation, FCFL \cite{pareto} proposed a two-stage optimization to solve a multi-objective optimization with fairness constraints, which demands more communication rounds. FPFL \cite{enforcing} utilized differential multipliers optimization method to solve main objective by taking into account fairness as a constraint, which is similar to this work. FedFB \cite{improving} adjusted the weight of the local loss function for each sensitive group during the aggregation phase. The objective formulation of FPFL, however, is not smooth, which may hinder the convergence of gradient-based learning. Moreover, the theoretical convergence guarantee is missing for the aforementioned works. 

\section{Preliminaries}
In this section, we introduce some mathematical notations and group fairness notions. After that, we briefly describe the framework of minimax FL.
\subsection{Notations}
Throughout this paper, we primarily focus on supervised binary classification tasks with binary sensitive attributes. The dataset is denoted as $\mathcal D = X \times Y \times S$ with size $|\mathcal D|$ constituting of an input image $X$, a label $Y = \{0,1\}$, and a sensitive attribute $S = \{0, 1\}$. We slightly abuse the notation of $\mathcal D$, $X$, $Y$, and $S$ to represent both the set and the distribution. The datasets can also be partitioned based on sensitive attributes, $\mathcal D^{s_0} = X \times Y \times S_0$ and $\mathcal D^{s_1} = X \times Y \times S_1$.

Some mathematical notations are stated as follows. $[N]$ denotes $\{1,2, ..., N\}$, $\|.\|$ represents the $\ell_2$-norm, and $\mathbbm{1}$ denotes the indicator function. We use $\mathcal W \subseteq \mathbb R^d$ and $\Lambda$ to represent the parameter spaces of the model $w$ and an additional learnable training parameter $\lambda$, respectively. Denote $f_w: X \rightarrow Y$ as a deep-learning model parameterized by $w$, taking $X$ as an input, and outputting the predicted label $Y$, and denote $q_w: X \rightarrow \mathbb R^2$ as the logits of the model, where the first element corresponds to $Y=0$ and the second element corresponds to $Y=1$.
\subsection{Group Fairness Metrics}
\label{sec:fairnessdefinitions}
To evaluate the group fairness performance of a machine learning model $f$, there are various notions based on how likely the model predicts a favorable outcome ($\hat Y = f(X) = 1$) for each group. \textit{Demographic parity} (DP) \cite{DP} is commonly used for assessing the fairness of the model. $f$ satisfies DP if the model prediction of favorable label is independent of $S$, i.e.,
\begin{IEEEeqnarray}{Rl}
    \mathbb E_{X|S=0}[f(X) = 1] = \mathbb E_{X|S=1}[f(X) = 1]. \label{eq:dp}
\end{IEEEeqnarray}
Another way to define the notion of group fairness is \textit{accuracy parity} (AP) \cite{disparatetreatment}. To satisfy this notion, $f$ conforms to the following equality
\begin{IEEEeqnarray}{Rl}
     \mathbb E_{\mathcal D^{s_0}} [\mathbbm{1}_{f(X) \neq Y}] = \mathbb E_{\mathcal D^{s_1}} [\mathbbm{1}_{f(X) \neq Y}]. \label{eq:dispmis}
\end{IEEEeqnarray}
In some use cases where the preference of users belonging to a sensitive group is considered, it is amenable to adopt \textit{equal opportunity} (EO) \cite{eo} of positive outcomes for each sensitive attribute as a fairness notion, mathematically written as
\begin{IEEEeqnarray}{Rl}
    \mathbb E_{X|S=0,Y=1}[f(X) = 1] =\mathbb E_{X|S=1,Y=1}[f(X) = 1].
\end{IEEEeqnarray}
In practice, it is difficult to achieve perfect fairness imposed by aforementioned fairness notions. To measure how close $f$ is to satisfy DP, we employ demographic parity difference metric $(\Delta_{DP})$ on favorable label, which is defined as
\begin{IEEEeqnarray}{Rl}
    \Delta_{DP} = |\mathbb E_{X|S = 0} [\mathbbm{1}_{f(X) = 1}] - \mathbb E_{X|S = 1} [\mathbbm{1}_{f(X) = 1}]|. \label{eq:dpp}
\end{IEEEeqnarray}
Similarly, the closeness measure to satisfy EO condition, equal opportunity difference ($\Delta_{EO}$) is defined as
\begin{align*}
    \Delta_{EO} = |\mathbb E_{X|Y = 1, S = 0} [\mathbbm{1}_{f(X) = 1}] - \mathbb E_{X|Y = 1, S = 1} [\mathbbm{1}_{f(X) = 1}]|. \numberthis
\end{align*}
These two closeness metrics are commonly used for assessing group fairness in machine learning models. Since only samples rather than the true data distribution are available, the metrics are estimated using the samples.

\begin{figure}
\centering
\includegraphics[width=0.45\textwidth]{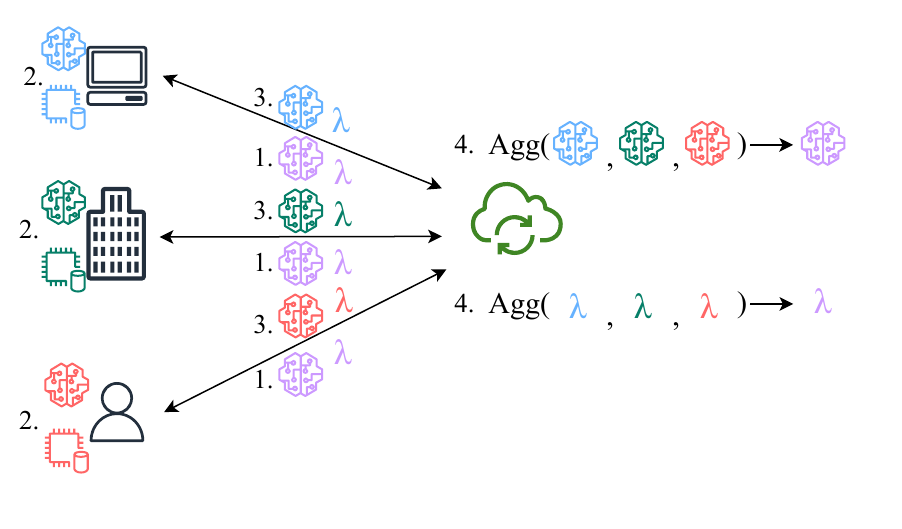}
\caption{The minimax FL framework. In each communication round, there are four steps in FL training: 1. Broadcasting phase 2. Local training phase 3. Client-to-server communication phase 4. Aggregation phase.} \label{federatedsetup}
\end{figure}

\subsection{Minimax FL Framework}
\label{subsec:flframework}
For a FL system with $N$ clients and one server, the goal is to train a global deep learning model $f_w$ on each client dataset $\mathcal D_i$ $(i \in [N])$ without sharing their datasets. In addition, an additional learning parameter (e.g. Lagrangian dual)  $\lambda \in \Lambda$ that aids the training can be exchanged between the server and clients, and processed on the clients and the server. Its generic procedure consists of four phases. Firstly, the clients receive the global model from the server (broadcasting phase) and train the model on their own dataset (local training phase). After that, the clients send their model update and the dual parameter update to the server (client-to-server communication phase), and the server aggregates the received updates from each participating client to get an updated model (aggregation phase). This process is repeated until convergence or a specified communication round, as illustrated in Fig. \ref{federatedsetup}.

During local training phase, each client aims to minimize their local risk function $F_i(w,\lambda)$ to update their model. At the same time, it maximizes the local risk function to update $\lambda$, in contrast to the conventional FL framework. For any regularization-based federated algorithms, the explicit formulation for the true local risk function of the $i-$th client represented by a loss function $l(f_w(x),y) = l(x,y;w)$ and a regularization function $g$ is given by
\begin{IEEEeqnarray}{Rl}
    F_i(w,\lambda):= \mathbb E_{(x_j, y_j) \sim \mathcal D_i}[l(x_j,y_j;w)] + g(\mathcal D_i;\lambda, w), 
\end{IEEEeqnarray}
and the corresponding empirical risk function is given by
\begin{IEEEeqnarray}{Rl}
    F_{i,S}(w,\lambda):= \frac{1}{|\mathcal D_i|} \sum_{(x_j,y_j) \sim \mathcal D_i}l(x_j,y_j;w) + g(\mathcal D_i;\lambda, w). 
\end{IEEEeqnarray}
The ultimate goal is to solve the following minimax objective of the true global risk function $F(w, \lambda)$
\begin{IEEEeqnarray}{Rl}
    \min_w \max_\lambda F(w, \lambda) := \sum_{i=1}^N p_i F_i(w, \lambda), 
\end{IEEEeqnarray}
where $p_i$ is the client coefficient with $\sum_{i=1}^N p_i = 1$ and $p_i \in [0,1]$. In FedAvg, the coefficient is set to the proportion of the samples from each client. Since the clients only have access to samples rather than the data distribution, the objective is replaced with the global empirical risk function defined as 
\begin{IEEEeqnarray}{Rl}
    F_S(w, \lambda) := \sum_{i=1}^N p_i F_{i,S}(w, \lambda).
\end{IEEEeqnarray}

\section{FFALM}
We first introduce the problem formulation for FL with group fairness constraints. Subsequently, we describe the proposed algorithm to achieve the objective. Lastly, we offer the convergence rate of the proposed algorithm.

\subsection{Solving Group Fairness Issue}
\subsubsection{Problem formulation} The objective of this work is to ensure group fairness on the FL-trained binary classification model. We tackle the problem by enforcing fairness during the local training. Specifically, the local training aims to minimize the local risk function while satisfying a notion of fairness. The strategy is to reformulate the local risk function as a sum of the main objective, which is related to the task performance, and the term related to fairness constraint, weighted with a learnable $\lambda$. In this way, we can use minimax FL framework as described in the previous section.

One of the essential requirements of minimax FL framework is to have a smooth local risk function. Since the indicator functions appearing in the fairness notions in Section \ref{sec:fairnessdefinitions} are not differentiable functions with respect to the model parameters, we need to replace all indicator functions with their corresponding surrogate continuous functions. In the case of AP, the choice of such continuous functions is readily available, which is cross entropy loss $CE(y,q_w(x)) = - \log \sigma(q_w(x)_y)$, where $\sigma(x) = \frac {1} {1+e^{-x}}$ is a sigmoid function. This is similar to how 0-1 loss ($\mathbb E_{\mathcal D}[\mathbbm{1}_{\hat Y \neq Y}]$) in the basic gradient-based learning can be replaced with cross-entropy loss \cite{surrogateloss}. The only difference in the formulation of AP is that it is conditioned on each sensitive attribute.  

To this end, we write the objective of the local training of the $i$-th client as a constrained optimization
\begin{IEEEeqnarray}{Rl}
    \min_{w} L_{S}(w,\mathcal D_i) \textrm{  s.t.  }  \mu(w,\mathcal D_i^{s_0}) = \mu(w,\mathcal D_i^{s_1}),  \label{eq:constrained}
\end{IEEEeqnarray}
where $L_{S}(w,\mathcal D) := \frac{1}{|\mathcal D|} \sum_{(x_j,y_j) \sim \mathcal D} l(x_j,y_j;w)$ and $\mu(w, \mathcal D) := \frac 1 {| \mathcal D|}\sum_{(x, y) \in \mathcal D} CE(y, q_w(x))$. It is worth mentioning that we also estimate (\ref{eq:dispmis}) from the samples, as shown in the definition of $\mu(w, \mathcal D)$.

\subsubsection{Local training phase} This problem can be approximately solved by following similar techniques from the augmented Lagrangian approach \cite{numericalopt} by treating the constraint as a soft constraint. Specifically, it seeks a saddle solution of an augmented Lagrangian function $\mathcal L_S$ parameterized by a suitable choice of penalty coefficient ($\beta$)
\begin{IEEEeqnarray}{Rl}
    \mathcal L_S(w_t, \lambda_{t-1}, \mathcal D_i) &:= L_S(w_t,\mathcal D_i) + \frac {\beta} 2\Delta_{\mu}(w_t, \mathcal D_i)^2 \IEEEnonumber\\
    &+ \lambda_{t-1} \Delta_{\mu}(w_t, \mathcal D_i), \label{eq:augmentedLagrangian} 
\end{IEEEeqnarray}
where $\Delta_{\mu}(w_t, \mathcal D_i) := \mu(w_t, \mathcal D_i^{s_0}) - \mu(w_t, \mathcal D_i^{s_1})$ by introducing a sub-optimizer $O$ to seek $w_t$ such that $\| \nabla_w \mathcal L_S(w_t, \beta, \mathcal D_i) \|$ is sufficiently small. Afterwards, $\lambda_{t-1}$ is updated to close the infeasibility gap, and the process is repeated. If the algorithm converges to the solution $(w^*, \lambda^*)$ of \eqref{eq:augmentedLagrangian} that satisfies second-order sufficient conditions \cite{numericalopt}, $w^*$ is the global solution to \eqref{eq:constrained}. 

Translating this view into FL, we can assign the sub-optimizer $O$ to the local training and the iteration index $t$ to the communication round. Hence, we can formulate the local training of $i$-th client as a two-stage process
\begin{IEEEeqnarray}{Rl}
    w_{i,t} &= \min_w  F_{i,S}(w,\lambda_{t-1}) = \min_w\mathcal L_S(w,\lambda_{t-1},\mathcal D_i)\label{eq:clientwupdate}\\
    \lambda_{i,t} &= \lambda_{t-1} + \eta_{\lambda,t} \Delta_{\mu}(w_t, \mathcal D_i) \label{eq:closefeasibility}.
\end{IEEEeqnarray}
Note that in the original augmented Lagrangian method, $\eta_{\lambda, t}$ is set to $\beta$. As shown later in the experiment results section, this proposed two-stage optimization gives more competitive results in terms of fairness performance.

Stochastic gradient descent is used to solve (\ref{eq:clientwupdate}), similar to FedAvg. Specifically, the $i$-th client computes the stochastic gradient of $\mathcal L^{(t,k)}_i = \mathcal L_S(w_i^{(t,k)}, \lambda_{i,t}, \mathcal B_i)$ at communication round $t$ and local iteration $k$ from its batch samples $\mathcal B_i$ sampled from its local distribution $\mathcal D_i$ as

\begin{IEEEeqnarray}{Rl}
    \nabla_w \mathcal L_i^{(t,k)} &= \nabla_w \biggl(L_S(w_i^{(t,k)},\mathcal B_i) + \lambda\Delta_{\mu}(w_i^{(t,k)}, \mathcal B_i) \IEEEnonumber \\
    &+ \frac {\beta} 2\Delta_{\mu}(w_i^{(t,k)}, \mathcal B_i)^2\biggr). 
    \label{eq:localgradw}
\end{IEEEeqnarray}

\subsubsection{Aggregation phase} The server receives model updates, $w_{i,t}$, as well as the dual updates, $\lambda_{i,t}$, from the clients. Following FedAvg, the received dual update from each client is aggregated by weighted average with the same client coefficient ($p_i$) as model aggregation
\begin{IEEEeqnarray}{Rl}
    w_{t} = \sum_{i=1}^N p_i w_{i,t} \quad\text{and}\quad \lambda_{t} = \sum_{i=1}^N p_i \lambda_{i,t}. \label{eq:ffalmaggregation}
\end{IEEEeqnarray}

The proposed algorithm is summarized in Algorithm \ref{alg:fairfedavgalm}.
\begin{algorithm}
    \caption{FFALM Algorithm}
    \label{alg:fairfedavgalm}
    \begin{algorithmic}[1]
    \State \textbf{Inputs:} $N$, $\{\mathcal D_i \}_{i=1}^N$, $\beta$, $\eta_{w,t}$, the number of local iteration $E$, and the maximum communication round $T$. 
    \State Randomly initialize the global model $w_0$ and set $\lambda_0 = 0$ on the server side
    \For{$t = 1$ to $T$}
        \State Broadcast $w_{t-1}$ and $\lambda_{t-1}$ to all clients
        \ForEach {$i \in [N] $}
        \State $w_i^{(t,0)} \gets  w_{t-1}$
        \For {$k = 1$ to $E$}
        \State Randomly sample the batch $\mathcal B_i$ from $ \mathcal D_i$
        \State Compute $\nabla_w \mathcal L_i ^{(t,k-1)}$ from (\ref{eq:localgradw})
        \State $w_i^{(t,k)} \gets w_i^{(t,k-1)} - \eta_{w,t} \nabla_w \mathcal L_i ^{(t,k-1)}$
        \EndFor
        \State Compute $\lambda_{i,t}$ from (\ref{eq:closefeasibility})
        \State Send $w_{i,t} = w_i^{(t,E)}$ and $\lambda_{i,t}$ to the server
        \EndFor
        \State Update $\eta_{w,t}$ using a LR scheduler
        \State Aggregation phase to obtain $w_t$ and $\lambda_t$ following (\ref{eq:ffalmaggregation})
    \EndFor
    \end{algorithmic}
\end{algorithm}
\subsection{Theoretical Convergence Guarantee}
\label{subsec:theory}
The proposed algorithm can be viewed as solving a minimax problem with $F_S(w,\lambda)=\sum_{i=1}^N p_i \mathcal L_S(w,\lambda_{t-1},\mathcal D_i)$. We provide the upper bound of the convergence rate based on how close the empirical primal risk function $R_S(w_t) := \max_\lambda  \mathcal L(w_t, \lambda)$ is to the optimal. Before presenting the result, we list several definitions and key assumptions. 
\begin{definition}
\label{assumption:smooth}
     Define a function $h: \mathcal W \times \Lambda \rightarrow \mathbb R$. $h(\cdot,\cdot)$ is $L$-smooth if it is continuously differentiable and there exists a constant $L > 0$ such that for any $w, w' \in \mathcal W$, $\lambda, \lambda' \in \mathbb R$, and $\xi \in \mathcal D$,
\begin{IEEEeqnarray}{Rl}
    \left\| \begin{pmatrix}
        \nabla_w h(w, \lambda; \xi) - \nabla_w h(w', \lambda'; \xi) \IEEEnonumber\\
        \nabla_\lambda h(w, \lambda; \xi) - \nabla_\lambda h(w', \lambda'; \xi)
    \end{pmatrix} \right\|
    \leq L \left\| \begin{pmatrix}
                w - w' \\
                \lambda - \lambda'
              \end{pmatrix}
              \right\|.
\end{IEEEeqnarray}
\end{definition} 
\begin{definition}
    \label{assumption:sc}
    $h(w,\cdot)$ is $\rho$-strongly convex if for all $w \in \mathcal W$ and $\lambda, \lambda' \in \Lambda$, $h(w,\lambda) \geq h(w, \lambda') + \langle\nabla_\lambda h(w,\lambda'),\lambda - \lambda'\rangle + \frac {\rho}{2} \|\lambda - \lambda'\|^2$.
\end{definition} 
\begin{definition}
\label{assumption:stronglyconcave}
    $h(w,\cdot)$ is $\rho$-strongly concave if $-h(w,\cdot)$ is $\rho$-strongly convex. 
\end{definition}
\begin{assumption}[Bounded Variance]
\label{assumption:boundedvar}
For randomly drawn batch samples $\xi$ and for all $i\in[N]$, the gradients $\nabla_w F_{i,S}(w, \lambda; \xi)$ and $\nabla_{\lambda} F_{i,S}(w, \lambda; \xi)$ have bounded variances $B_w$ and $B_\lambda$ respectively. If $g_{i,w}(w, \lambda|\xi) := \nabla_w F_{i,S}(w,\lambda;\xi)$ is the unbiased local estimator of the gradient, $\mathbb E_\xi [\|g_{i,w}(w, \lambda|\xi) - \nabla_w F_{i,S}(w, \lambda)\|^2] \leq B_w^2$, and the case for $\lambda$ is similar but bounded by $B_\lambda^2$. 
\end{assumption}

\begin{assumption}[Bounded Dissimilarity]
\label{assumption:boundedgrad}
    There exist $\beta \geq 1$ and $\kappa \geq 0$ such that
    \begin{align}
        \sum_{i=1}^N p_i\|\nabla_w F_{i,S}(w,\lambda)\|^2 \leq \beta^2\| \sum_{i=1}^N p_i\nabla_w F_{i,S}(w,\lambda)\|^2 + \kappa^2.
    \end{align}
\end{assumption}
\noindent
\begin{assumption}[Bounded Gradient]
\label{assumption:lipschitz}
    The gradient of the local risk functions are bounded in norm by $G$,
    \begin{align}
        \|\nabla_w F_{i,S}(w,\lambda)\| \leq G.
    \end{align}
\end{assumption}
\noindent
\begin{definition}[]
    A function $h(\cdot,\lambda)$ satisfies the PL condition if for all $\lambda$, there exists a constant $\mu > 0$ such that, for any $w \in \mathcal W$, $\frac 1 2 \|\nabla h(w)\|^2 \geq \mu(h(w) - \min_{w'\in \mathcal W} h(w'))$.
\end{definition}
\noindent
Definition \ref{assumption:smooth}, Assumption \ref{assumption:boundedvar} and Assumption \ref{assumption:boundedgrad} are commonly used for federated learning \cite{wang2021field}. Assumption \ref{assumption:boundedgrad} is satisfied when the gradient clipping method is employed. Lastly, the PL-condition of $\mathcal L(\cdot,\lambda)$ is shown to hold on a large class of neural networks \cite{losslandscape}.

For simplicity, we assume full participation and the same number of local iterations for each client. The minimum empirical primal risk is $R^*_S = \min_w R_S(w)$. The upper bound of the convergence rate of FFALM is given by the following theorem. 
\begin{thm}
Define $\kappa = \frac L {\mu}$. Let $\eta_{w,t} = \frac {2\beta^2} {\mu t}$ and $\eta_{\lambda, t} = \mathcal O(\frac{1}{t^{2/3}})$. Given that Assumption \ref{assumption:boundedvar}, Assumption \ref{assumption:boundedgrad}, and Assumption \ref{assumption:lipschitz} hold, each $F_{i,S}(w,\lambda)$ is $L$-smooth, each $F_{i,S}(\cdot,\lambda)$ satisfies $\mu$-PL condition, and each $F_{i,S}(w,\cdot)$ is $\rho$-strongly concave, we have
\begin{align*}
    \mathbb E R_S(w_{T+1}) - R_S^* = \mathcal O \biggl(\frac{\Gamma + B_w^2  + B_\lambda^2 }{T^{2/3}} \biggr),
\end{align*}
after $T$ communication rounds, where $\Gamma := F_S^* - \sum_{i=1}^N p_i F_{i,S}^*$, $F_S^* := \min_w \max_\lambda F_S(w,\lambda)$ and  $F_{i,S}^* := \min_w \max_\lambda F_{i,S}(w,\lambda)$. 
\label{thm:convergencerate}
\end{thm}
\begin{IEEEproof}
   See Appendix \ref{appendix}.
\end{IEEEproof}
\noindent
Note that $\Gamma$ quantifies statistical heterogeneity of the FL system. In the case of strong non-iid, the saddle solution of the global risk function is significantly different from the weighted sum of each saddle local risks. 

\section{Empirical Study}
\begin{figure*}
\centering
\begin{subfigure}[b]{0.5\textwidth}
   \includegraphics[width=1\linewidth]{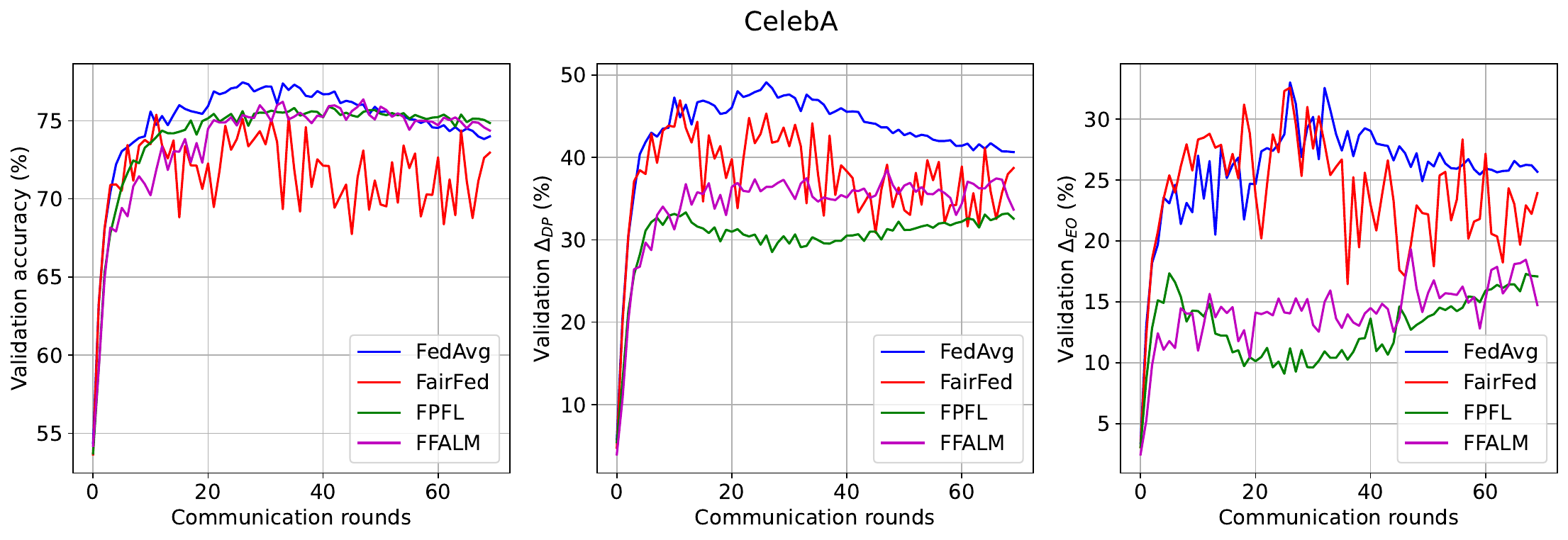}
   \caption{}
   \label{fig:celeba-curve}
\end{subfigure}\hfill
\begin{subfigure}[b]{0.5\textwidth}
   \includegraphics[width=1\linewidth]{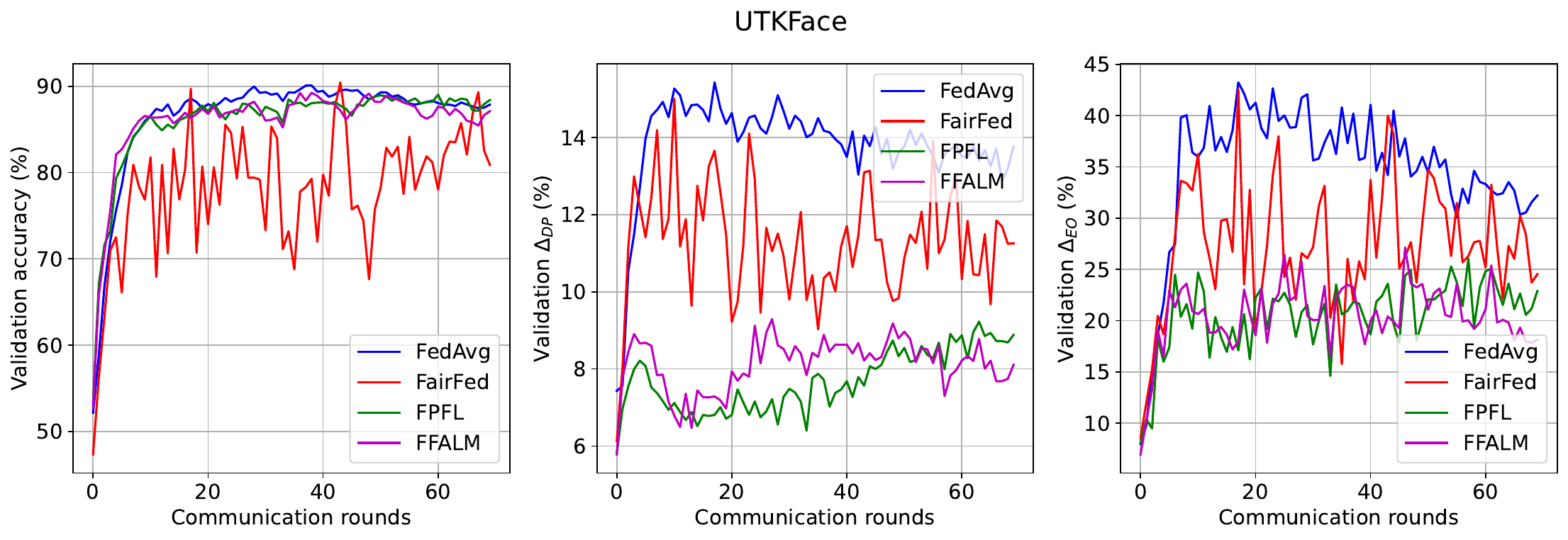}
   \caption{}
   \label{fig:utkface-curve}
\end{subfigure}
\caption{Learning curves on validation set for (a) CelebA dataset and (b) UTKFace dataset}
\label{fig:trainingcurves}
\end{figure*}
In this section, we evaluate the effectiveness of FFALM based on three important performance metrics: the prediction accuracy, DPD, and EOD on real-world datasets. We provide the results and comparison with other existing FL algorithms. 

\subsection{Datasets and Tasks} We want to investigate how the classification model trained in FL, which is designed to further increase the prediction accuracy on specific domain, influences the fairness performance. In particular, two datasets from CV domain are used in this study: CelebA \cite{celeba}, and UTKFace \cite{utkface}, and ResNet-18 \cite{resnet} models are used for both datasets. The task of CelebA dataset is a binary classification for predicting attractiveness in images with gender as the sensitive attribute, and the task of UTKFace dataset is to determine whether the age of a person from an image is above or below 20 with skin color as the sensitive attribute. The favorable labels for CelebA and UTKFace are attractive and age below 20 respectively. Each dataset is split into three categories: training set, validation set, and testing set.

\subsection{FL Setting} There are 10 clients participating in FL training. We synthetically simulate statistical heterogeneity by introducing label skews, which can be implemented using Dirichlet distribution parameterized by $\alpha$ on the proportion of samples for a given class and client on a given centralized samples \cite{niid}. The case of severe data heterogeneity is investigated in this experiment by setting $\alpha=0.3$. The experiment for each instance is repeated 10 times with different seeds. The FL training ends after 70 communication rounds.

\subsection{Baselines} The following are the baselines used for the comparison study.
\begin{enumerate}
    \item \textbf{FedAvg.} It is the universal baseline in FL which trains the model locally without considering fairness and aggregates all model updates by weighted average. 
    \item \textbf{FairFed \cite{fairfed}.} The server receives the local DP metrics, and based on them and the global trend, the server adjusts the value of $p_i$ adaptively before averaging the model updates.
    \item \textbf{FPFL  \cite{enforcing}.} It enforces fairness by solving the constrained optimization on the sample loss function $L_S$ with two constraints. These constraints ensure that the absolute difference between the overall loss and the loss of each sensitive group ($\delta_{\mu_0,i} = [L(w^t, \mathcal D_i) -  \mu(w^t, \mathcal D_i^{s_0})]_+$ and $\delta_{\mu_1,i} = [L(w^t, \mathcal D_i) -  \mu(w^t, \mathcal D_i^{s_1})]_+$, where $[x]_+ := \max(0,x)$) does not deviate from a specific threshold. We set this threshold to be zero. Hence, we reformulate it as a local constrained optimization with
    \begin{IEEEeqnarray}{Rl}
        g(\mathcal D_i; \lambda, w) &= \lambda_0 \delta_{\mu_0,i}+ \lambda_1 \delta_{\mu_1,i}
        + \frac {\beta} 2 (\delta_{\mu_0,i}^2 +\delta_{\mu_1,i}^2 ). \IEEEnonumber\\ 
        \end{IEEEeqnarray}
\end{enumerate}

\label{sec:hyperparameter}
\begin{table}[htpb]
\centering
\caption{Comprehensive list of hyperparameter values used in the experiments on CelebA and UTKFace datasets for baselines and FFALM.}
\label{table:hyper} 
\begin{tabular}{ p{3.0cm} p{2.5cm} p{0.7cm} }
  \toprule
    Hyperparameters           & Algorithms                 & Values       \\ 
  \midrule
    Batch size & all                    & 128 \\
    Gradient clipping on $w$     & all & 1.0                       \\
    LR of $w$ decay step size    & all & 50                        \\
    LR or $w$ decay step factor              & all               & 0.5                      \\
    $\eta_{w,0}$                 & all                          & 0.05                      \\ 
    $\beta$ & FairFed & 0.5 \\
    $\beta$ & FPFL & 5.0 \\
    $\eta_{\lambda, t}$ & FPFL &    0.5 \\
    $b$ & FFALM & 1.05 \\
    $\beta$ & FFALM &  2.0 \\
    $\eta_{\lambda,0}$ & FFALM & 2.0 \\
    $\lambda_0$ & FFALM and FPFL & 0.0 \\
  \bottomrule
  \label{table:hyperparameters}
\end{tabular} 
\end{table}

\subsection{Implementation Details}
The hyperparameters used in the experiment for FFALM and all baselines are shown in Table \ref{table:hyperparameters}. Following the setup from augmented Lagrangian method, we slowly increase the learning rate of $\lambda$ by a factor $b$ per communication round for FFALM. All algorithms have the same learning rate scheduler of $w$ which is realized by a constant-step decay factor.

\begin{table}[hbtp]
    \centering
    \caption{Comparison of the performance of the proposed algorithm across baselines on CelebA and UTKFace datasets. $\uparrow$ indicates the larger the value the better and $\downarrow$ indicates the smaller the value the better.}
    \resizebox{\columnwidth}{!}{
    \sisetup{detect-weight=true,detect-inline-weight=math,tight-spacing = true,
         table-format = 2.2,}
        \begin{tabular}{l
        S[table-format=2.2]
        S[table-format=2.2]
        S[table-format=2.2]
        }
            \toprule
            {Algorithm} &
           {Acc $\uparrow(\%)$} & {$\Delta_{DP}$ $\downarrow(\%)$} & {$\Delta_{EO}$ $\downarrow(\%)$} \\
            \midrule
            \multicolumn{4}{c}{CelebA} \\
            \midrule
            FedAvg & \bfseries  74.66 & 39.54 & 21.11 \\
            FairFed & 73.27 & 37.36 & 19.50 \\
            FPFL & 74.58 & 30.87 & 9.71 \\
            FFALM & 74.06 & \bfseries 28.92 & \bfseries 6.82  \\
            \midrule
            \multicolumn{4}{c}{UTKFace} \\
            \midrule
            FedAvg & 86.22 & 13.07 & 19.12 \\
            FairFed & 79.54 & 10.80 & 17.92 \\
            FPFL & \bfseries 86.73 &  7.97 & 15.85 \\
            FFALM & 86.02 & \bfseries 6.60 & \bfseries 13.87 \\
            \bottomrule
        \end{tabular}
    }
    \label{table:result}
\end{table}

\subsection{Results}
The training curves of all algorithms are shown in Figure \ref{fig:trainingcurves}. The curves show how accuracy, DPD, and EOD change as the communication round increases. It can be observed that FFALM shows similar improvement in accuracy as FedAvg while improving the fairness performance.

The experimental results evaluated on the testing set are presented in Table \ref{table:result}. For celebA dataset, FFALM improves DPD by almost $11 \%$ and EOD by roughly $14 \%$ compared to FedAvg. FFALM outperforms other baselines in fairness performance with minimal accuracy loss. For UTKFace dataset, FFALM improves DP difference performance by about $6 \%$, and reduces EO gap by $6 \%$ compared to FedAvg. The overall fairness improvement is also apparent on FFALM in UTKFace dataset compared with different baselines.

\section{Conclusion}
In this paper, we proposed FFALM, an FL algorithm based on augmented Lagrangian framework to handle group fairness issues. It leveraged accuracy parity constraint for smooth loss
formulation of minimax FL framework. It was shown that the theoretical convergence rate of FFALM is $\mathcal O(\frac 1 {T^{2/3}})$. Experiment results on CelebA and UTKFace datasets demonstrated the effectiveness of the proposed algorithm in improving fairness with negligible accuracy drop under severe statistical heterogeneity.

\appendices

\section{Proof of Theorem 1}
\label{appendix}
\label{app:convergence}
We introduce and prove several lemmas before proving Theorem 4.2. The proof is similar to Theorem 3 in \cite{dpsgda} except we have additional lemmas to cover the local training phases and statistical heterogeneity that exist in FL.
\begin{lemma}[\cite{lemma7}]
    Assume $L$-smoothness for $F_{i,S}$ with $i \in [N]$ and $F_{i,S}(w,\cdot)$ is $\rho$-strongly concave. Let $\lambda$ be bounded. Then the function $R_S(w)$ is $L + \frac {L^2} \rho$-smooth and $\nabla R_S(w) = \nabla_w F_S(w, \hat \lambda_S(w))$, where $\hat \lambda_S(w) = \arg\max_\lambda F_S(w, \lambda)$. Moreover, $\hat \lambda_S(w)$ is $\frac L \rho$-Lipschitz continuous.
\label{lemma:surrogate}
\end{lemma}
\begin{lemma}
    Assume $F_{i,S}$ with $i \in [N]$ is $L$-smooth and satisfies and all $F_{i,S}(w,\cdot)$ with $i \in [N]$ is $\rho$-strongly concave. Assume that the FL system data dissimilarity can be bounded, \linebreak
    i.e. $\sum_{i=1}^N p_i \|\nabla_w F_{i,S}(w,\lambda)\|^2 \leq \beta^2\| \sum_{i=1}^N p_i \nabla_w F_{i,S}(w,\lambda)\|^2 + \kappa^2 $. With $\Gamma := F^*_S - \sum_{i=1}^N p_i F^*_{i,S}$, where $F_S^* := \min_w \max_\lambda F_S(w,\lambda)$ and  $F_{i,S}^* := \min_w \max_\lambda F_{i,S}(w,\lambda)$, we have 
    \begin{IEEEeqnarray}{Rl}
        \|\nabla R_S(w) \|^2 \geq \frac{2\mu}{\beta^2}(R_S(w) - \min_w R_S(w) + \Gamma) -\frac{\kappa^2}{\beta^2}.
    \end{IEEEeqnarray}
\label{lemma:PL}
\end{lemma} 
\begin{IEEEproof}
    \begin{align*}
        &\| \nabla_w F_S(w, \hat \lambda_S(w)) \|^2 = \|\sum_{i=1}^N p_i \nabla_w F_{i,S}(w,\hat \lambda_S(w))\|^2 \nonumber\\
        &\geq  \frac{\sum_{i=1}^N p_i \| \nabla_w F_{i,S}(w,\hat \lambda_S(w)) \|^2 - \kappa^2}{\beta^2} \nonumber\\
        &\geq \frac{\sum_{i=1}^N 2\mu p_i(F_{i,S}(w,\hat \lambda_S(w)) -  \min_{w' \in \mathcal W} F_{i,S}(w',\hat \lambda_S(w)))- \kappa^2}{\beta^2} \nonumber\\
        &\geq \frac{\sum_{i=1}^N 2\mu p_i(F_{i,S}(w,\hat \lambda_S(w)) -  \min_{w' \in \mathcal W} F_{i,S}(w',\hat \lambda_S(w')))- \kappa^2}{\beta^2} \nonumber\\
        &= \frac{2\mu(F_S(w,\hat \lambda_S(w)) - \sum_{i=1}^N p_i \min_{w' \in \mathcal W} F_{i,S}(w',\hat \lambda_S(w')))- \kappa^2}{\beta^2}  \nonumber\\
        &= \frac{2\mu}{\beta^2}(F_S(w,\hat \lambda_S(w)) - \min_w F_S(w, \hat \lambda_S(w)) + \Gamma) -\frac{\kappa^2}{\beta^2}.
    \end{align*}
    The first inequality is obtained by Jensen inequality, and the second inequality comes from the $L$-smoothness property. Since $\nabla_w R_S(w) = \nabla_w F_S(w, \hat \lambda_S(w))$ by Lemma \ref{lemma:surrogate}, the proof is complete.
    
\end{IEEEproof}
\begin{lemma}
Assume Assumption \ref{assumption:boundedgrad} holds for each $F_{i,S}$ and each $F_{i,S}$ with $i \in [N]$ is $L$-smooth. If the number of local iterations is $E$, we have
\begin{IEEEeqnarray}{Rl}
      &\mathbb E\|\nabla R_S(w_t) - \sum_{i=1}^N p_i\sum_{k=1}^{E}\nabla_wF_{i,S}(w_i^{(t,k)}, \lambda_t)\|^2 \leq EL^2\mathbb E [\||\lambda^*_S(w_t)-\lambda_t \|^2] \nonumber\\&+ \eta_{w,t}^2L^2\frac{E(E+1)}{2}(B_w^2 + G^2) .
\end{IEEEeqnarray}
\label{lemma:c5}
\end{lemma}

\begin{IEEEproof} 
We expand the definition of $R_S(w_t)$ and apply Jensen's inequality to get
\begin{align}
     &\mathbb E\|\nabla R_S(w_t) - \sum_{i=1}^N p_i\sum_{k=1}^{E}\nabla_wF_{i,S}(w_i^{(t,k)}, \lambda_t)\|^2  \nonumber\\
    &\leq \mathbb E\sum_{i=1}^N p_i\sum_{k=1}^{E}  [\|\nabla_w F_{i,S}(w_i^{(t,0)}, \lambda^*_S(w_t)) \nonumber\\&- \nabla_w F_{i,S}(w_i^{(t,k)}, \lambda_t) \|^2] \nonumber\\
    &\leq \mathbb E \sum_{i=1}^N p_iL^2\sum_{k=1}^{E}([\||\lambda^*_S(w_t)-\lambda_t \|^2] + [\|w^{(t,0)}-w_i^{(t,k)} \|^2]). \nonumber\\
     &= \mathbb E L^2E[\|\lambda^*(w_t)-\lambda_t \|^2] \nonumber\\&+ \eta_{w,t}^2L^2\sum_{i=1}^N p_i \sum_{k=1}^E\mathbb E\|\sum_{k'=1}^k g_{i,w}(w_i^{(t,k')}, \lambda_t; \xi_i^{(t,k')}) \|^2\nonumber\\
     &\leq L^2E\mathbb E\|\lambda^*_S(w_t)-\lambda_t \|^2 \nonumber\\&+ \eta_{w,t}^2L^2\sum_{i=1}^N p_i\sum_{k=1}^{E}\sum_{k'=1}^k\mathbb E\| g_{i,w}(w_i^{(t,k')}, \lambda_t; \xi_i^{(t,k')}) \|^2 \nonumber\\
     &\leq L^2E\mathbb E\|\lambda^*_S(w_t)-\lambda_t \|^2 + \eta_{w,t}^2L^2\sum_{i=1}^N p_i\sum_{k=1}^{E}\sum_{k'=1}^k (B_w^2 + G^2) \nonumber\\
     &\leq L^2E\mathbb E [\||\hat \lambda^*_S(w_t)-\lambda_t \|^2] + \eta_{w,t}^2L^2\frac{E(E+1)}{2}(B_w^2 + G^2). \label{eq:wts}
\end{align}
The second inequality follows from the fact that the function is $L$-smooth. We derive the first equality by considering the $k$-th local update on the $i$-th client. The third inequality is another application of Jensen's inequality. The fourth inequality is a consequence of Assumption \ref{assumption:boundedvar}, which bounds the variance of the gradient estimators and Assumption \ref{assumption:lipschitz}.
\end{IEEEproof}

\begin{lemma}
    Rewrite the combined primal update (local update + aggregation) of the global model at round $t$ ($w_t$) as
    $w_{t+1} = w_t - \eta_{w,t}\sum_{i=1}^N p_i \sum_{k=1}^{E}g_{i,w}(w^{(t,k)}_i,  \lambda_t;\xi_i^{(t,k)})$, where $E$ is the number of local iterations. Assume that Assumption \ref{assumption:boundedvar} and the conditions from Lemma \ref{lemma:c5}, \ref{lemma:PL}, and \ref{lemma:surrogate} hold. We have
\begin{align}
    \mathbb E [R_S(w_{t+1}) - R^*_S] &\leq (1-\frac{\mu}{\beta^2} \eta_{w,t})(\mathbb E R_S(w_t) - R^*_S) \nonumber\\&+\frac{EL^2\eta_{w,t}} 2\mathbb E[\||\lambda^*_S(w_t)-\lambda_t \|^2]\nonumber\\
    & + \frac{(L + \frac {L^2} \rho)\eta_{w,t}^2} 2 E^2B_w^2 +\frac{1}{2\beta^2} \eta_{w,t}\kappa^2 \nonumber\\&-\frac{\mu}{\beta^2} \eta_{w,t}\Gamma + \frac{\eta_{w,t}^3L^2E(E+1)}{4}(B_w^2 + G^2). 
\end{align}
\label{lemma:wterm}
\end{lemma}
\begin{IEEEproof} 
We start from the smoothness of $R_S$ and choose two points, $w_{t+1}$ and $w_t$. 
\begin{align}
    \mathbb ER_S(w_{t+1}) - R^*_S &\leq \mathbb E\Big[ R_S(w_t) -  R^*_S  + \langle \nabla_w R_S(w_t), w_{t+1} - w_t \rangle \nonumber\\&+ \frac{L + \frac {L^2} \rho} 2 \|w_{t+1} - w_t\|^2\Big] \nonumber\\
    &=\mathbb E\Big[ R_S(w_t) -  R^*_S  \nonumber\\&- \eta_{w,t} \langle \nabla_w R_S(w_t), \sum_{i=1}^N p_i \sum_{k=1}^{E} g_{i,w}(w^{(t,k)}_i,  \lambda_t;\xi_i^{(t,k)}) \rangle \nonumber\\
    &+ \frac{(L + \frac {L^2} \rho)\eta_{w,t}^2} 2 \left \|\sum_{i=1}^N p_i \sum_{k=1}^{E} g_{i,w}(w^{(t,k)}_i,  \lambda_t;\xi_i^{(t,k)}) \right \|^2\Big] \nonumber\\
    &=\mathbb E R_S(w_t) -  R^*_S - \nonumber\\&\eta_{w,t}\mathbb E \langle \nabla_w R_S(w_t), \sum_{i=1}^N p_i \sum_{k=1}^{E} \nabla F_{i,S}(w^{(t, k)}, \lambda_t)  \rangle \nonumber\\
    &+ \frac{(L + \frac {L^2} \rho)\eta_{w,t}^2} 2  \mathbb E\|\sum_{i=1}^N p_i \sum_{k=1}^{E} g_{i,w}(w^{(t,k)}_i,  \lambda_t;\xi_i^{(t,k)}) \nonumber\\&- \sum_{i=1}^N p_i\sum_{k=1}^{E} \nabla F_{i,S}(w^{(t,k)}_i, \lambda_t) \nonumber\\
    &+ \sum_{i=1}^N p_i \sum_{k=1}^{E} \nabla F_{i,S}(w^{(t,k)}_i, \lambda_t)\|^2 \nonumber\\
    &\leq \mathbb E R_S(w_t) -  R^*_S \nonumber\\&- \eta_{w,t} \mathbb E\langle \nabla_w R_S(w_t), \sum_{i=1}^N p_i\sum_{k=1}^{E} \nabla F_{i,S}(w_i^{(t, k)}, \lambda_t) \rangle \nonumber\\
    &+ \frac{(L + \frac {L^2} \rho)\eta_{w,t}^2} 2 \mathbb E\left\| \sum_{i=1}^N p_i \sum_{k=1}^{E} \nabla F_{i,S}(w^{(t,k)}_i, \lambda_t) \right\|^2 \nonumber\\&+ \frac{(L + \frac {L^2} \rho)\eta_{w,t}^2} 2 E^2B_w^2. 
\end{align}
The first equality arises from the combined primal update. We obtain the second equality by invoking Assumption \ref{assumption:boundedvar}, which relies on the unbiasedness of the gradient estimators. The final inequality follows from the fact that we can zero out uncorrelated gradient estimators between distinct clients or local iterations, apply Jensen's inequality, and subsequently apply the following inequality\\ 
$\mathbb E \left \|\sum_{i=1}^N p_i \sum_{k=1}^{E} ( g_{w}(w^{(t,k)}_i,  \lambda_t;\xi_i^{(t,k)}) - \nabla F_{i,S}(w^{(t,k)}_i, \lambda_t)) \right \|^2 \leq E^2B_w^2$. Further simplifying the term with $\eta_{w,t} \leq \frac 1 {(L + L^2/\rho)}$, and using Lemma \ref{lemma:PL}, we obtain
\begin{align}
    \mathbb E [R_S(w_{t+1}) - R^*_S] &\leq \mathbb E R_S(w_t) - R^*_S \nonumber\\&+ \frac{\eta_{w,t}} 2 \mathbb E \|\nabla_w R_S(w_t) - \sum_{i=1}^N p_i\sum_{k=1}^E \nabla F_{i,S}(w^{(t, k)}, \lambda_t) \|^2 \nonumber\\
    &-\frac{\eta_{w,t}} 2\mathbb E \|\nabla_w R_S(w_t) \|^2\ + \frac{(L + \frac {L^2} \rho)\eta_{w,t}^2} 2 E^2B_w^2\nonumber\\
    &\leq(1-\frac{\mu}{\beta^2}\eta_{w,t})(\mathbb E R_S(w_t) - R^*_S) \nonumber\\&+ \frac{\eta_{w,t}}{2}\mathbb E\|\nabla_w R_S(w_t) - \sum_{i=1}^N p_i\sum_{k=1}^{E} \nabla F_{i,S}(w^{(t, k)}, \lambda_t)\|^2 \nonumber\\&
    + \frac{(L + \frac {L^2} \rho)\eta_{w,t}^2} 2 E^2B_w^2  - \frac{\mu}{\beta^2} \eta_{w,t} \Gamma  +\frac{\eta_{w,t}}{2\beta^2} \kappa^2
    \nonumber\\&\leq (1-\frac{\mu}{\beta^2} \eta_{w,t})(\mathbb E R_S(w_t) - R^*_S) \nonumber\\&+\frac{EL^2\eta_{w,t}} 2\mathbb E\||\lambda^*_S(w_t)-\lambda_t \|^2\nonumber\\
    & + \frac{(L + \frac {L^2} \rho)\eta_{w,t}^2} 2 E^2B_w^2 +\frac{\eta_{w,t}}{\beta^2}\kappa^2 \nonumber\\&-\frac{\mu}{\beta^2} \eta_{w,t}\Gamma + \frac{\eta_{w,t}^3L^2E(E+1)}{4}(B_w^2 + G^2), 
\end{align}
where the second inequality is obtained from Lemma \ref{lemma:PL}, and the third inequality is obtained from Lemma \ref{lemma:c5}.
\end{IEEEproof}
\begin{lemma}
     Rewrite the combined dual update as $\lambda_{t+1} = \lambda_t + \eta_{\lambda, t}\sum_{i=1}^N p_i g_{i,\lambda}(w^{(t,k)}_i,  \lambda_t;\xi_i^{(t,k)})$. Assuming the conditions from Lemma \ref{lemma:wterm} and Assumption \ref{assumption:boundedvar} hold, we have
    \begin{align}
         \mathbb E \|\lambda_{t+1} - \lambda^*_S(w_{t+1})\|^2 &\leq (1 + \frac 1 {\epsilon})\frac{L^2\eta_{w,t}^2E}{\rho^2}(B_w^2 + G^2)
        \nonumber\\&+ (1+\epsilon)(1-\rho \eta_{\lambda,t})\|\lambda_t - \lambda^*_S(w_t)\|^2 \nonumber\\
        &+ (1+\epsilon)\eta_{\lambda,t}^2B_\lambda^2
    \end{align}
    for any $\epsilon > 0$.
\label{lemma:vterm}
\end{lemma} 
\begin{IEEEproof}
     By Young's inequality, for any $\epsilon > 0$, we have
    \begin{align}
        \|\lambda_{t+1} - \lambda^*_S(w_{t+1})\|^2 &\leq (1+\epsilon)\|\lambda_{t+1} - \lambda^*_S(w_{t})\|^2 \nonumber\\&+ (1 + \frac 1 \epsilon)\|\lambda^*_S(w_{t}) - \lambda^*_S(w_{t+1})\|^2. \label{young}
    \end{align}
    For the second term, using the fact that $\lambda^*_S(w)$ is $\frac L \rho$-Lipschitz (Lemma \ref{lemma:surrogate}) and applying the expectation to get
    \begin{align}
        &\mathbb E[\|\lambda^*_S(w_{t+1}) - \lambda^*_S(w_t) \|^2] \leq \frac{L^2}{\rho^2} \mathbb E[\|w_{t+1} - w_t\|^2] \nonumber\\&= \frac{L^2\eta_{w,t}^2}{\rho^2}\mathbb E[\|\sum_{i=1}^N p_i\sum_{k=1}^{E} g_{w}(w^{(t,k)}_i,  \lambda_t;\xi_i^{(t,k)})\|^2] \nonumber\\
        &=  \frac{L^2\eta_{w,t}^2}{\rho^2} \mathbb E\bigl\| \sum_{i=1}^N p_i\sum_{k=1}^{E} g_{w,i}(w^{(t,k)}_i,  \lambda_t;\xi_i^{(t,k)})- \nabla_w F_{i,S}(w^{(t,k)}_i, \lambda_t)) \nonumber\\
        &+   \sum_{i=1}^N p_i\sum_{k=1}^{E} \nabla_w F_{i,S}(w^{(t,k)}_i, \lambda_t) \|^2  \nonumber\\
        &\leq \frac{L^2\eta_{w,t}^2}{\rho^2}(EB_w^2 + \bigl\|\sum_{i=1}^N p_i\sum_{k=1}^{E} \nabla_w F_{i,S}(w^{(t,k)}_i, \lambda_t)\|^2) \nonumber\\
         &\leq \frac{L^2\eta_{w,t}^2}{\rho^2}(EB_w^2 + \sum_{i=1}^N p_i\sum_{k=1}^{E}\bigl\| \nabla_w F_{i,S}(w^{(t,k)}_i, \lambda_t)\|^2) \nonumber\\
         &\leq \frac{L^2\eta_{w,t}^2E}{\rho^2}(B_w^2 + G^2). \nonumber\\
        \label{youngone}
    \end{align}
    The first equality stems from the combined primal update. We derive the second inequality from Assumption \ref{assumption:boundedvar}. The third inequality is a consequence of Jensen's inequality, while the fourth inequality follows from Assumption \ref{assumption:lipschitz}.
    
    For the first term, applying the combined dual update, we get
    \begin{align}
        &\mathbb E [\|\lambda_{t+1} - \lambda^*_S(w_{t})\|^2] \leq \mathbb E [\|\lambda_t + \eta_{\lambda,t}\sum_{i=1}^N p_i g_{i,\lambda}(w^{(t,E)}_i,  \lambda_t;\xi_i^{(t,k)}) \nonumber\\&-  \lambda^*_S(w_t)\|^2] \nonumber\\
        &\leq \mathbb E\|\lambda_t - \lambda^*_S(w_t) \|^2 + 2\eta_{\lambda,t}\mathbb E[\langle \lambda_t - \lambda^*_S(w_t), \nonumber\\&\sum_{i=1}^N p_i  g_{i,\lambda}(w^{(t,E)}_i,  \lambda_t;\xi_i^{(t,k)}) \rangle] \nonumber\\
        &+  \eta_{\lambda,t}^2 \mathbb E [\|\sum_{i=1}^N p_i  (g_{i,\lambda}(w^{(t,E)}_i,  \lambda_t;\xi_i^{(t,k)})-\nabla_\lambda F_{i,S}(w_i^{(t,E)},\lambda_t)) \nonumber\\&+ \nabla_\lambda F_{i,S}(w_i^{(t,E)},\lambda_t) \|^2]\nonumber\\
        &\leq  \mathbb E\|\lambda_t - \lambda^*_S(w_t) \|^2 + 2\eta_{\lambda,t}\langle \lambda_t - \lambda^*_S(w_t), \sum_{i=1}^N p_i  \nabla_\lambda F_{i,S}(w^{(t,E)}, \lambda_t) \rangle\nonumber\\
        &+ \eta_{\lambda,t}^2\sum_{i=1}^N p_i\|\nabla_\lambda F_S(w_i^{(t,E)}, \lambda_t) \|^2 + \eta_{\lambda,t}^2B_\lambda^2 \nonumber\\
        &\leq (1-\rho\eta_{\lambda,t})\mathbb E\|\lambda_t - \lambda^*_S(w_t) \|^2 \nonumber\\
        &+2\eta_{\lambda,t}\sum_i p_i (F_{i,S}(w_i^{(t,E)}, \lambda_t) - F_{i,S}(w_i^{(t,E)}, \lambda^*_S(w_t))) \nonumber \\
        &+ \eta_{\lambda,t}^2\sum_{i=1}^N p_i\|\nabla_\lambda F_S(w_i^{(t,E)}, \lambda_t) \|^2 + \eta_{\lambda,t}^2B_\lambda^2 \nonumber\\
        &\leq  (1-\rho\eta_{\lambda,t})\mathbb E\|\lambda_t - \lambda^*_S(w_t) \|^2 - \frac{\eta_{\lambda,t}}{L}\sum_i p_i\|\nabla_\lambda F_{i,S}(w_i^{(t,E)}, \lambda_t)\|^2 \nonumber\\
        &+  \eta_{\lambda,t}^2\sum_{i=1}^N p_i\|\nabla_\lambda F_S(w_i^{(t,E)}, \lambda_t) \|^2+ \eta_{\lambda,t}^2B_\lambda^2 \nonumber\\
        &\leq (1-\rho\eta_{\lambda,t})\mathbb E\|\lambda_t - \lambda^*_S(w_t) \|^2+ \eta_{\lambda,t}^2B_\lambda^2.
       \label{youngtwo}
    \end{align}
    We obtain the third inequality by invoking Assumption \ref{assumption:boundedvar}. The fourth inequality arises from the fact that $F_{i,S}(w,\cdot)$ is $\rho$-strongly concave. The fifth inequality is obtained from the property of \ref{assumption:smooth}. The final inequality follows from the bound $\eta_{\lambda,t} \leq \frac 1 L$. By combining equations \eqref{youngone} and \eqref{youngtwo} with equation \eqref{young}, we arrive at the desired result.
\end{IEEEproof}
\begin{lemma}
    Define $a_t = \mathbb E [R_S(w_{t}) - R^*_S]$ and $b_t =  \mathbb E \|\lambda_{t} - \lambda^*_S(w_{t})\|^2$. For any non-increasing sequence $\{ \nu_t > 0\}$ and any $\epsilon > 0$, we have the following relations,
\begin{align}
    &a_{t+1} + \nu_{t+1}b_{t+1} \leq k_{1,t} a_t + k_{2,t} \nu_t b_t + c_t, 
\end{align}
where
\begin{align}
    k_{1,t} &= 1- \frac{\mu \eta_{w,t}}{\beta^2}  \\
    k_{2,t} &= \frac{EL^2\eta_{w,t}} {2\nu_{t}}+ (1 + \epsilon)(1-\rho\eta_{\lambda,t}) \\
    c_t &=  - \frac{\mu}{\beta^2} \eta_{w,t}\Gamma +\frac{\eta_{w,t}}{2\beta^2} \kappa^2+ \frac{(L + \frac {L^2} \rho)\eta_{w,t}^2} 2 E^2B_w^2  \nonumber\\&+ \frac{\eta_{w,t}^3L^2E(E+1)}{4}(B_w^2 + G^2)  \nonumber\\
        &+\nu_{t}(1 + \frac 1 {\epsilon})\frac{L^2E\eta_{w,t}^2}{\rho^2}(B_w^2+G^2)+ \nu_{t}(1+\epsilon)\eta_{\lambda,t}^2B_\lambda^2.
\end{align}
\label{lemma:atbt}
\end{lemma}
\begin{IEEEproof}
    The result follows from combining Lemma \ref{lemma:wterm} and Lemma \ref{lemma:vterm}.
\end{IEEEproof}
We can proceed to prove Theorem \ref{thm:convergencerate}. From Lemma \ref{lemma:atbt}, by choosing $\epsilon = \frac{\rho\eta_{\lambda,t}}{2(1-\rho\eta_{\lambda,t})}$, we have
\begin{align*}
    k_{2,t} = \frac{EL^2\eta_{w,t}} {2\nu_t} + 1 - \frac{\rho\eta_{\lambda,t}}{2}.
\end{align*}
Then, set $\nu_t = \frac{2L^2\eta_{w,t}}{\rho\eta_{\lambda,t}} $. we have
\begin{align}
     k_{1,t} &= 1- \frac{\mu\eta_{w,t}}{\beta^2}\\
    k_{2,t} &= 1 - \frac{\rho \eta_{\lambda,t}} 4.
\end{align}
By setting $\eta_{\lambda,t}\geq \frac{2\mu}{\rho\beta^2}\eta_{w,t}$, we get
\begin{align}
     &a_{t+1} + \nu_{t+1}b_{t+1} \leq (1 -\frac \mu {2\beta^2} \eta_{w,t})(a_t + \nu_t b_t) - \frac{\mu}{\beta^2} \eta_{w,t}\Gamma +\frac{\eta_{w,t}}{2\beta^2} \kappa^2 \nonumber\\&+ \frac{(L + \frac {L^2} \rho)\eta_{w,t}^2} 2 E^2B_w^2  + \frac{\eta_{w,t}^3L^2E(E+1)}{4}(B_w^2 + G^2)  \nonumber\\
        &+\nu_{t}(1 + \frac 1 {\epsilon})\frac{L^2E\eta_{w,t}^2}{\rho^2}(B_w^2+G^2)+ \nu_{t}(1+\epsilon)\eta_{\lambda,t}^2B_\lambda^2.  \label{eq:thm2intermediate}
\end{align}
Choose $\eta_{w,t} = \frac {2\beta^2} {\mu t}$ and $\eta_{\lambda, t} = \min\{\frac{1}{L}, \frac 4 {\rho t^{2/3}}\}$ and multiply both sides of (\ref{eq:thm2intermediate}) by $t$ to get
\begin{align}
    t(a_{t+1} + \nu_{t+1}b_{t+1}) &\leq (t-1)(a_t + \nu_t b_t)+ \frac{4\beta^4E^2B_w^2(L+\frac{L^2}{\rho})}{\mu^2 t}\nonumber\\&+ \frac{L^2E(E+1)8\beta^6}{\mu^3t^2}(B_w^2 + G^2) \nonumber\\
    &+ \frac{2L^4E\beta^6}{\rho^2\mu^3t^{2/3}}+\frac{4\beta^2L^2}{\mu\rho^2t^{2/3}} + \frac{4\beta^2L^2}{\mu\rho^2(t^{2/3}-4)} \nonumber\\&+ \frac{\kappa^2}{\mu}- \frac{\Gamma}{2}.
\end{align}
Applying the inequality inductively  from $t=0$ to $T$ and divide both sides by $T$, we get
\begin{align}
    \mathbb E[R_S(w_{T+1}) - R^*_S] &\leq \frac{4\beta^4E^2B_w^2(L+\frac{L^2}{\rho})\ln T}{\mu^2 T} \nonumber\\&+ \frac{L^2E(E+1)8\beta^6}{\mu^3T^2}(B_w^2 + G^2) + \frac{6L^4E\beta^6}{\rho^2\mu^3T^{2/3}} \nonumber\\ &+ \frac{12\beta^2L^2}{\mu\rho^2T^{2/3}} + \frac{12\beta^2L^2}{\mu\rho^2(T^{2/3}-4)} \nonumber\\ &+\frac{\kappa^2}{\mu}- \frac{\Gamma}{2}.
\end{align}\IEEEQED
\section{Analysis of The Upper Bound of The Convergence Rate on a Simple Example}
The theoretical result shows that on nonconvex setting, the algorithm could converge to a different (non-optimal) saddle point. This can be understood from a simple example consisting of two clients with the following local risk functions
\begin{align}
    F_{1}(w,\lambda) &= (w-1)^2 + 2sin^2(w-2) - (\lambda-4)^2 \nonumber\\
    F_{2}(w,\lambda) &= (w+1)^2 + 2.1sin^2(w+2) - (\lambda+4)^2.
\end{align}
Also, define the global risk function as $F_S(w,\lambda) = 0.5 F_1(w,\lambda) + 0.5 F_2(w,\lambda)$. It can be shown that $F_1^* \approx 0.657645$,  $F_2^* \approx 0.669149$, and $(w^*, \lambda^*) \approx (0.650928, 0)$ with $F^* \approx -13.3915$. There exists another saddle point where it satisfies the second-derivative test at $(w,\lambda) \approx (-0.633757, 0)$ with $F(w,\lambda) = -13.3552$. It can also be shown that by restricting the domain, $w \in [-3, 3]$ and $\lambda \in [-5,5]$, we have $\mu = \frac 1 {8}, \rho=1, L = 18, G=11, \kappa^2 = 3.7^2, \beta^2 = 1, \Gamma \approx -14.72$. This means that the nonvanishing term has a nonzero positive dependency with the heterogeneity parameters. If we run FFALM with $(w_0, \lambda_0) = (2,2)$ for 1 million communication rounds, we get $(w_T, \lambda_T) \approx (-0.6338911, 0)$. This shows that the algorithm may converge to different saddle point locations.
\section{Additional Experiments on Different Levels of Heterogeneity}
\begin{table}[hbtp]
    \centering
    \caption{Comparison of the performance of the proposed algorithm across baselines on CelebA and UTKFace datasets on other levels of heterogeneity ($\alpha=1.0$ and $\alpha=5.0$). $\uparrow$ indicates the larger the value the better and $\downarrow$ indicates the smaller the value the better.}
    \resizebox{\columnwidth}{!}{
    \sisetup{detect-weight=true,detect-inline-weight=math,tight-spacing = true,
         table-format = 2.2,}
        \begin{tabular}{l
        S[table-format=2.2]
        S[table-format=2.2]
        S[table-format=2.2]
        }
            \toprule
            {Algorithm} &
           {Acc $\uparrow(\%)$} & {$\Delta_{DP}$ $\downarrow(\%)$} & {$\Delta_{EO}$ $\downarrow(\%)$} \\
            \midrule
            \multicolumn{4}{c}{CelebA ($\alpha=1.0$)} \\
            \midrule
            FedAvg & 76.76 & 41.53 & 22.66 \\
            FairFed & 75.36 & 39.37 & 21.99 \\
            FPFL & \bfseries 77.54 & 36.41 & 16.94  \\
            FFALM  & 77.47 & \bfseries 35.00 & \bfseries 14.00 \\
            \midrule
            \multicolumn{4}{c}{CelebA ($\alpha=5.0$)} \\
            \midrule
            FedAvg & 77.36 & 41.78 & 22.79 \\
            FairFed & 77.1 &  \bfseries 41.40 & \bfseries 22.24 \\
            FPFL & \bfseries 79.28 & 42.99 & 22.48 \\
            FFALM & 78.89 & 41.55 & 23.08 \\
            \midrule
            \midrule
            \multicolumn{4}{c}{UTKFace ($\alpha=1.0$)} \\
            \midrule
            FedAvg & \bfseries 91.95 & 15.59 & 15.94 \\
            FairFed & 89.14 & 13.52 & 13.23 \\
            FPFL & 87.56 & \bfseries 11.08 & 4.38  \\
            FFALM & 90.39 & 13.89 & \bfseries 3.87 \\
            \midrule
            \multicolumn{4}{c}{UTKFace ($\alpha=5.0$)} \\
            \midrule
            FedAvg & \bfseries 92.02 & 15.74 & 9.35 \\
            FairFed & 91.25 &  15.5 & 10.09 \\
            FPFL & 86.74 & 17.38 & \bfseries 7.37 \\
            FFALM & 88.48 & \bfseries 11.78 & 8.43 \\
            \bottomrule
        \end{tabular}
    }
    \label{table:result}
\end{table}
\newpage
\bibliography{ref}
\bibliographystyle{ieeetr}

\end{document}